\documentclass[sigconf]{acmart}

\AtBeginDocument{%
  }

\copyrightyear{2025}
\acmYear{2025}
\setcopyright{acmlicensed}

\acmConference[MUWS '25] {Proceedings of the 4th International Workshop on Multimodal Human Understanding for the Web and Social Median}{October 27--31, 2025}{Dublin, Ireland.}
\acmBooktitle{Proceedings of the 4th International Workshop on Multimodal Human Understanding for the Web and Social Media (MUWS '25), October 27--31, 2025, Dublin, Ireland}
\acmISBN{979-8-4007-1838-0/2025/10}
\acmDOI{10.1145/XXXXXX.XXXXXX}

\usepackage{makecell}
\usepackage{soul}
\usepackage{multirow}
\usepackage{libertine}
\usepackage{graphicx}
\graphicspath{{./figures/}}

\begin{document}

\title{Landmark Guided Visual Feature Extractor for Visual Speech Recognition with Limited Resource}

\author{Lei Yang}
\affiliation{%
 \institution{Shanghai Jiao Tong University}
 \city{Shanghai}
 \country{China}
}
\email{yangleisx@sjtu.edu.cn}
\orcid{0000-0002-7508-2341}

\author{Junshan Jin}
\affiliation{%
 \institution{Shanghai Jiao Tong University}
 \city{Shanghai}
 \country{China}
}
\email{jjs030212@sjtu.edu.cn}

\author{Mingyuan Zhang}
\affiliation{%
 \institution{Shanghai Jiao Tong University}
 \city{Shanghai}
 \country{China}
}
\email{zmy_00@sjtu.edu.cn}

\author{Yi He}
\affiliation{%
 \institution{Shanghai Jiao Tong University}
 \city{Shanghai}
 \country{China}
}
\email{heyi96@sjtu.edu.cn}

\author{Bofan Chen}
\affiliation{%
 \institution{Shanghai Jiao Tong University}
 \city{Shanghai}
 \country{China}
}
\email{bofan822@sjtu.edu.cn}

\author{Shilin Wang}
\affiliation{%
 \institution{Shanghai Jiao Tong University}
 \city{Shanghai}
 \country{China}
}
\authornote{
  Corresponding author.
}
\email{wsl@sjtu.edu.cn}

\renewcommand{\shortauthors}{Yang et al.}

\begin{abstract}
  Visual speech recognition is a technique to identify spoken content in silent speech videos, which has raised significant attention in recent years.
  Advancements in data-driven deep learning methods have significantly improved both the speed and accuracy of recognition.
  However, these deep learning methods can be effected by visual disturbances, such as lightning conditions, skin texture and other user-specific features. Data-driven approaches could reduce the performance degradation caused by these visual disturbances using models pretrained on large-scale datasets.
  But these methods often require large amounts of training data and computational resources, making them costly.
  To reduce the influence of user-specific features and enhance performance with limited data, this paper proposed a landmark guided visual feature extractor. Facial landmarks are used as auxiliary information to aid in training the visual feature extractor. 
  A spatio-temporal multi-graph convolutional network is designed to fully exploit the spatial locations and spatio-temporal features of facial landmarks. Additionally, a multi-level lip dynamic fusion framework is introduced to combine the spatio-temporal features of the landmarks with the visual features extracted from the raw video frames.
  Experimental results show that this approach performs well with limited data and also improves the model's accuracy on unseen speakers.
\end{abstract}

\ccsdesc[500]{Computing methodologies~Computer vision}
\ccsdesc[500]{Computing methodologies~Speech recognition}

\keywords{Visual Speech Recognition, Facial Landmark, Spatio-Temporal Graph Convolutional Network}


\maketitle

\section{Introduction}
\label{sec:intro}
Visual speech recognition (VSR), also known as lip-reading, is a technique that recognize the spoken content based solely on visual information from silent videos. It can be used in aiding people with hearing impairments and in tasks such as film dubbing and speech recognition in noisy environments. It has received widespread attention from both academic and industrial communities\cite{ma2022chinese, ma2020transformer,he2024speaker,zhang2020can,cheema2024identification}.

Traditional visual speech recognition methods are often based on hand-crafted features, such as Deep Boltzmann Machine(DBM)\cite{ngiam2011multimodal}, Active Shape Model(ASM)\cite{wang2004real}, Active Appearance Model(AAM)\cite{papandreou2007multimodal,papandreou2009adaptive}, Spline model\cite{leung2006automatic}. These methods typically involve two stages: feature extraction and classification. The first stage extracts visual features from the input video frames, while the second stage employs classifiers such as Support Vector Machines (SVM)\cite{ngiam2011multimodal} or Hidden Markov Models (HMM)\cite{wang2004real,papandreou2007multimodal,lucey2007patch,papandreou2009adaptive} to recognize the spoken content. However, these traditional methods often struggle to capture complex temporal dependencies and spatial relationships in the data, leading to limited performance.

In recent years, end-to-end visual speech recognition systems based on deep neural networks have been developed. 
Existing deep learning based visual speech recognition algorithms typically consist of two stages: visual feature extraction and temporal feature modeling. In the first stage, visual features are extracted from input facial or lip image sequences cropped from raw videos, using 3D convolutional neural networks (CNNs) or time-distributed 2D CNNs. 
In the second stage, the architecture is tailored to the recognition task. For instance, sequence classification models are employed for word-level recognition, while sequence-to-sequence models are utilized for sentence-level tasks. Various techniques such as Temporal Convolution Network(TCN)\cite{Martinez2020, Ma2021}, Recurrent Neural network(RNN)\cite{Stafylakis2017, Petridis2017} are adopted to aggregate the temporal features and generate prediction. Attention mechanism has also been widely used\cite{Chen2021, Koumparoulis2022, burchi2023audio, chang2024conformer} due to their powerful sequence modeling capabilities.

However, end-to-end deep learning based methods that rely on visual cues are often susceptible to various visual perturbations, such as complex lightning conditions and skin texture, leading to performance degradation. This brings two primary challenges. Firstly, enhancing the model's robustness to such perturbations typically requires training with large-scale datasets, which incurs significant costs, including expenses related to data collection and the computational overhead associated with training large models. Secondly, in data-constrained scenarios, these models tend to exhibit poor generalization, particularly with unseen speakers. While training a personalized model for each target user may improve performance, it similarly entails substantial computational costs.

Large-scale datasets such as LRS2\cite{son2017lip} and LRS3\cite{afouras2018deep} have contributed to improvements in the performance of visual speech recognition models. Existing end-to-end approaches focus on recognition performance include: collection and annotation of large-scale real datasets\cite{chen2023cn, chang2024conformer}; using pseudo-labels generated from Automatic Speech Recognition (ASR) methods with audio for VSR training\cite{ma2023auto}; training user adaptation models\cite{he2024speaker}; using synthetic speech data for training\cite{liu2023synthvsr}, and the application of techniques such as knowledge distillation to leverage knowledge from ASR models\cite{ma2021lira, ma2022visual}.

Due to the fact that most existing datasets are in English, the performance of VSR in other languages with limited data still exhibits notable disparities. To some extent, this can be addressed through techniques like distillation across different languages, as seen in \cite{kim2023lip} and \cite{kondo2024inter}, or by employing pseudo-labeling methods\cite{yeo2024visual}. However, the development of end-to-end visual speech recognition methods primarily relies on the availability of extensive training data and advancements in computational power.

Another challenge is the degradation in performance of VSR models when applied to unseen speakers. This issue arises because the models are typically trained on videos collected from a fixed set of speakers, meaning that the visual features they've learned during training may not generalize well to new speakers. Variations in lip shapes, skin tones, and other user-specific traits can significantly affect the performance of VSR models.

To reduce the influence of user-specific features and improve performance with limited data, auxiliary information can be incorporated to assist in model training. For instance, facial landmarks or facial parsing segmentation\cite{yang2023fine} can be used as supplementary information to enhance the model's robustness against visual disturbances.
The facial landmarks extracted by a face detector, pretrained over large scale face datasets\cite{yang2016wider, bulat2017far}, are relatively insensitive to changes in illumination and texture, providing more stable features and thereby enhancing the performance of visual speech recognition.
Since the preprocessing pipeline of visual speech recognition systems typically includes face detection, localization and landmark detection, the supplementary information from facial landmarks can be seamlessly integrated into subsequent processing without introducing extra computational costs.
In previous works\cite{liu2020lip, Xue2023, Li2024, Wu2024}, spatial position of facial landmarks are introduced to improve the lipreading performance of unseen speakers. In \cite{Li2024}, Spatio-Temporal Graph Convolutional Network (ST-GCN) is adopted to extract lip moving information from lip contour landmarks, which could reduce the influence of static characteristics such as lip appearance. In \cite{Wu2024}, Intra-Frame and Inter-Frame features got extracted and fused, and mutual information regularization are utilized to learn user-independent features. However, approaches that rely solely on either full-face or lip contour landmarks often struggle to capture rich spatio-temporal information to recognize the spoken content. Techniques using GCNs with fixed 1-hop adjacency matrices lack sufficient information, such as subtle changes in lip region, which is crucial for accurate visual speech recognition.

In this paper, a landmark guided visual feature extractor is proposed to further address these limitations and enhance recognition performance in data-limited scenarios, This method takes into account both the landmark coordinates and the dynamic features of the landmarks, using adjacency matrices based on spatial distance and similarity to extract more comprehensive information and enhance recognition performance. A spatio-temporal multi-graph convolutional modele is designed to fully exploit the spatial locations and spatio-temporal features of facial landmarks. Additionally, a multi-level lip dynamic fusion framework is introduced to combine the spatio-temporal features of the landmarks with the visual features extracted from the raw video frames. 

The key contributions of this paper are summarized as follows:
\begin{itemize}
  \item A new landmark guided visual feature extractor is proposed to enhance the performance of visual speech recognition with limited data. The facial landmarks feature, such as landmark coordinates and spatio-temporal features, are integrated to provide more comprehensive information for visual speaker recognition.
  \item A spatio-temporal multi-graph convolutional module is designed to fully exploit the spatial locations and subtle movements around the lip region. Three distinct graphs are constructed from the extracted spatio-temporal features and facial landmarks: a Landmark Coordinate Graph (LCG), a Distance-Aware Graph (DAG), and a Similarity-Aware Graph (SAG). The adjacency matrices of these graphs are constructed using fully connected adjacent matrices based on spatial distance and similarity, allowing for the extraction of rich information that  improves lipreading performance.
  \item The spatio-temporal features of the landmarks are fused with the visual features extracted from the original frames through a multi-level lip dynamic fusion framework. 
  \item Extensive experimental results on the word-level lipreading dataset LRW-ID and LRW demonstrate that the proposed methods can achieve commendable classification performance, particularly in scenarios with limited data availability.
\end{itemize}

\section{Challenges and Motivations}

\subsection{Challenges in Visual Speech Recognition}

Visual speech recognition models face several challenges in real-world scenarios, such as visual speaker authentication with random prompts, in-cabin authentication for intelligent vehicles, and assisting speech recognition in noisy environments.

A key issue is the complex lighting conditions often present in real-world scenarios, which can significantly affect the performance of models that depend on visual features. 
For example, variations in lighting could blur the lip contour, making it harder for models to recognize it accurately. Furthermore, complex textures in the lip region can interfere with vision-based models, preventing them from accurately locating the lip region and consequently hindering their ability to correctly interpret lip movements\cite{yang2023fine}.

Another challenge is that training visual speech recognition models often requires a large amount of annotated data, which is challenging to obtain in real-world scenarios.
In particular, the front-end visual feature extractor typically relies on a large-scale facial images to effectively capture the subtle spatiotemporal variations, which are essential for lipreading tasks.

In scenarios with limited training data, distribution shifts in attributes such as race and gender within the dataset can adversely affect the performance of visual feature extractors, making it challenging to train visual speech recognition models that could be well-generalized across different speakers. While developing user-adaptive models\cite{he2024speaker} or feature disentangled models\cite{he2024lip} could enhance the recognition performance of specific speakers, this also introduces additional training costs. To mitigate the influence of user-specific features and improve performance in data-limited condition, facial landmarks are introduced as auxiliary information to improve the model's robustness.

\subsection{Motivations}

\begin{figure}[!t]
  \centering
  \includegraphics[width=0.8\linewidth]{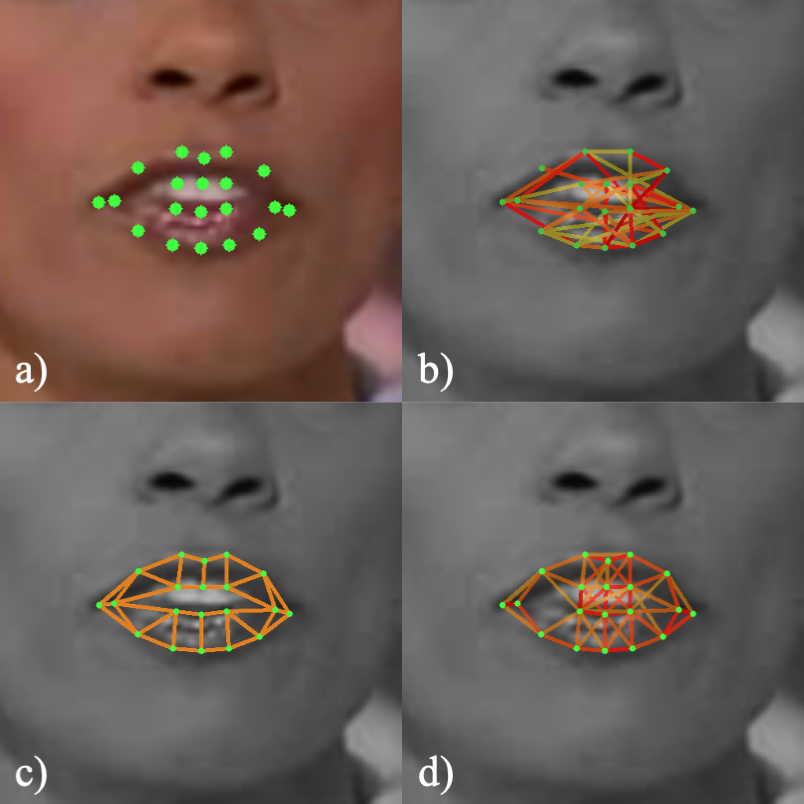}
  \caption{Three distinct graph are constructed based on the extracted spatio-temporal features and facial landmarks. a) the 20 facial landmarks around the lip contour; b) the adjacency matrix of the similarity-aware graph; c) the adjacency matrix of the landmark coordinate graph; d) the adjacency matrix of the distance-aware graph. The color scheme represents the link weights, with links of low weight omitted in the figure. Best view in color.}
  \label{fig:demo}
\end{figure}

In order to enhance training efficiency with limited datasets and improve detection performance on unseen speakers, facial landmark features are introduced as supplementary information. Those facial landmarks, extracted through a pretrained face detector, are robust to illumination, texture and other user-specific traits, providing more stable and reliable information.

Compared to models that utilize dual-stream networks, extracting rich information from landmark dynamics across multiple perspectives offers a more comprehensive approach.

Firstly, the positional information of the lips is crucial for recognition.
Based on the coordinates of 20 landmarks in the lip region, the shape and size of the lips can be determined, providing insights into the spoken content. Relative positional changes across  different areas also carry meaningful information, especially the positional relationships between adjacent key points, which offer rich detailed information about the lip region. Therefore, as shown in Fig.~\ref{fig:demo}, we have constructed a landmark coordinate graph convolution to fully utilize the coordinate information. This approach employs spatio-temporal graph convolution to comprehensively consider spatial positional relationships and temporal dynamic changes.

Secondly, we incorporate the local dynamic features of each landmark. These local dynamic features are extracted using 3D convolution.
We analyze the relationships between different landmarks based on distance and similarity.
Those landmark with higher similarity exhibit similar dynamic transformations, aiding the model in capturing significant lip movements. Landmarks that are closer in spatial distance also tend to have correlation. 
The adjacency weights are shown in Fig.~\ref{fig:demo}, where links with low weights are omitted for clarity. This approach captures dynamic features across multiple dimensions.

For the features extracted from different branches, a multi-level progressive fusion approach is adopted. Unlike directly merging visual features and landmark features, this method fully considers the differences between various perspectives, preventing potential conflicts and further improving the model's performance.

\section{The Proposed Method}
\label{sec:method}

\subsection{Overall Structure}
\label{ssec:structure}

\begin{figure*}[!t]
    \centering
    \includegraphics[width=.95\linewidth]{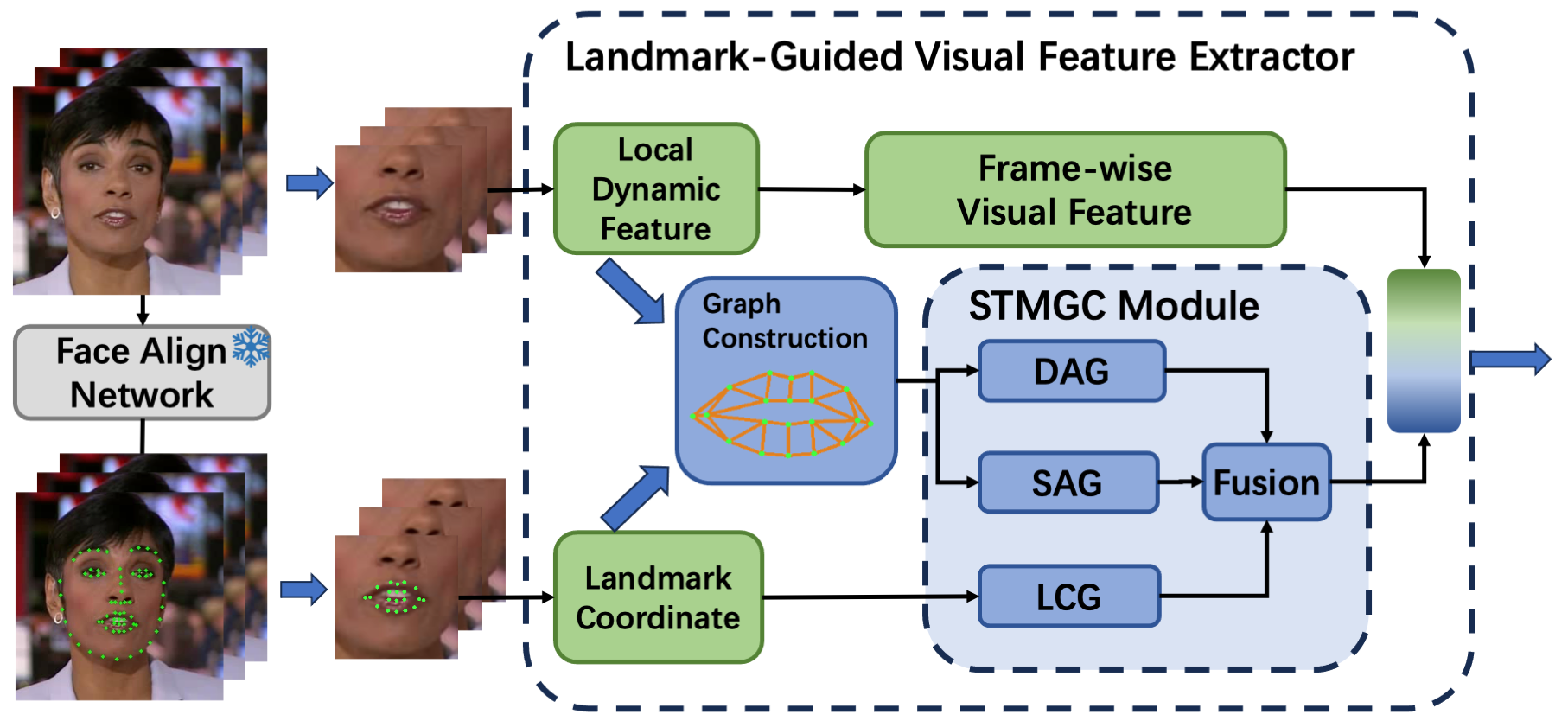}
    \caption{The overall architecture of the proposed network. The extracted spatio-temporal features and the landmarks are employed to construct three distinct landmarks graph.}
    \label{fig:overall}
\end{figure*}

The overall visual speech recognition model, as shown in Fig.~\ref{fig:overall}, follows the structure of previous works\cite{ma2022training}, consisting of a front-end landmark-guided visual feature extractor, a back-end temporal feature aggregator, and a classification head with temporal pooling. 

The front-end model includes a local dynamic feature extractor, a pretrained face align network to extract facial landmarks\cite{yang2016wider,bulat2017far}, a time-distributed frame-wise visual feature extractor based on ResNet18, and the proposed spatio-temporal multi-graph convolutional module which extract rich information from temporal features and landmark features. 

The features extracted by the front-end model are fed into a back-end network based on DC-TCN\cite{Ma2021}, and the resulting video-level features are passed through a classification head to obtain word-level recognition results.

\subsection{Visual Feature Extraction}
\label{ssec:vfe}

The visual feature extraction includes the initial local dynamic feature extraction \(\textbf{E}_{d}\), and frame-wise feature extraction, \(\textbf{E}_{v}\). 
\begin{align}
    {\bf F}_{d} &= \textbf{E}_{d}({\bf Frames})\\
    {\bf F}_{v} &= \textbf{E}_{v}({\bf F}_{d})
\end{align}

Local dynamic feature extraction \(\textbf{E}_{d}\)is implemented using 3D convolution, which captures local spatiotemporal dynamic features. Previous studies have demonstrated that initial 3D convolutions can effectively capture inter-frame relationships, thereby improving the recognition performance. The local dynamic features ${\bf F}_{d}$ obtained through 3D convolution maintain the spatiotemporal resolution without reduction while preserving global spatial structure.
When constructing graph convolutions using local dynamic features ${\bf F}_{d}$, the features of each graph node inherently contain spatiotemporal dynamic information, which facilitates the perception of complex spatiotemporal variation patterns through spatiotemporal graph convolutions.

Frame-wise feature extraction  \(\textbf{E}_{v}\) is implemented using ResNet18 with shared weights across the temporal sequence. This approach not only effectively extracts multi-level visual features but also reduces the number of parameters and computational cost through parameter sharing, thereby accelerating inference speed.

\subsection{Spatio-Temporal Multi-Graph Convolutional Module}
\label{ssec:mgcn}

In spatio-temporal multi-graph convolutional module, three landmark graphs are constructed, namely the Distance-Aware Graph (DAG), Landmark Coordinate Graph (LCG), and Similarity-Aware Graph (SAG).
The construction of these graphs is based on local dynamic feature ${\bf F}_{d}$. Each node in the landmark graph represents a facial landmark in the video, while the edges between nodes indicate the spatio-temporal relationships among these landmarks. 
In this paper, only 20 landmarks of lip contour are considered, as shown in Fig~\ref{fig:demo} a). For each landmark, we select the nearest spatial position and extract a vector of length C as the node feature.
By leveraging the local dynamic feature, the model can assign rich spatio-temporal features to each facial landmark, enabling the landmark graph to not only have high spatial resolution but also convey continuous dynamic information across the temporal dimension.
Different landamrk graphs are processed through graph convolution to extract global dynamic features. Finally, these features are fused with visual features through a multi-level fusion approach, serving as the basis for temporal feature aggregation and classification in the back end.

\subsubsection{Graph Construction}
\label{sssec:gc}

\paragraph{Distance Aware Graph}:
The Distance Aware Graph (DAG) contains 20 landmarks around the lip contour, and the node features are composed of spatio-temporal features extracted by the front-end visual feature extractor. The adjacency matrix is a fully connected matrix with weights, where the weight between nodes is inversely proportional to the L2 distance between them, thereby emphasizing the features of spatially adjacent nodes during the graph convolution process, which is demonstrated in bottom-right panel in Fig.~\ref{fig:demo}.

\paragraph{Landmark Coordinate Graph}:
A Landmark Coordinate Graph (LCG) is an unweighted graph that consists of 20 landmarks along the lip contour as demonstrated in top-left in Fig.~\ref{fig:demo}. The features of the nodes in the graph are the coordinates of these  facial landmarks in the original image, while the adjacency matrix represents the connections between each node and its one-hop neighbors along the lip contour, as well as the self-loop. The degree of each node is between 3 and 5, as shown in bottom-left panel in Fig.~\ref{fig:demo}.

\paragraph{Similarity Aware Graph(SAG)}: 
The spatio-temporal features are also utilized to construct the node features in a Similarity-Aware Graph (SAG). The construction of the adjacency matrix is based on a fully connected graph with weighted edges, where the weight between different nodes is proportional to the cosine similarity between their features. Considering that nodes with higher similarity are more likely to belong to the same semantic category, each node in SAG focus more on the features of similar nodes during the graph convolution process. As the bottom-right panel in Fig.~\ref{fig:demo}, nodes around the outer contour of the lips have higher similarity weights.

\subsubsection{Spatial-Temporal Graph Convolution Network}
\label{sssec:stgcn}

\begin{figure}[!t]
    \centering
    \includegraphics[width=.9\linewidth]{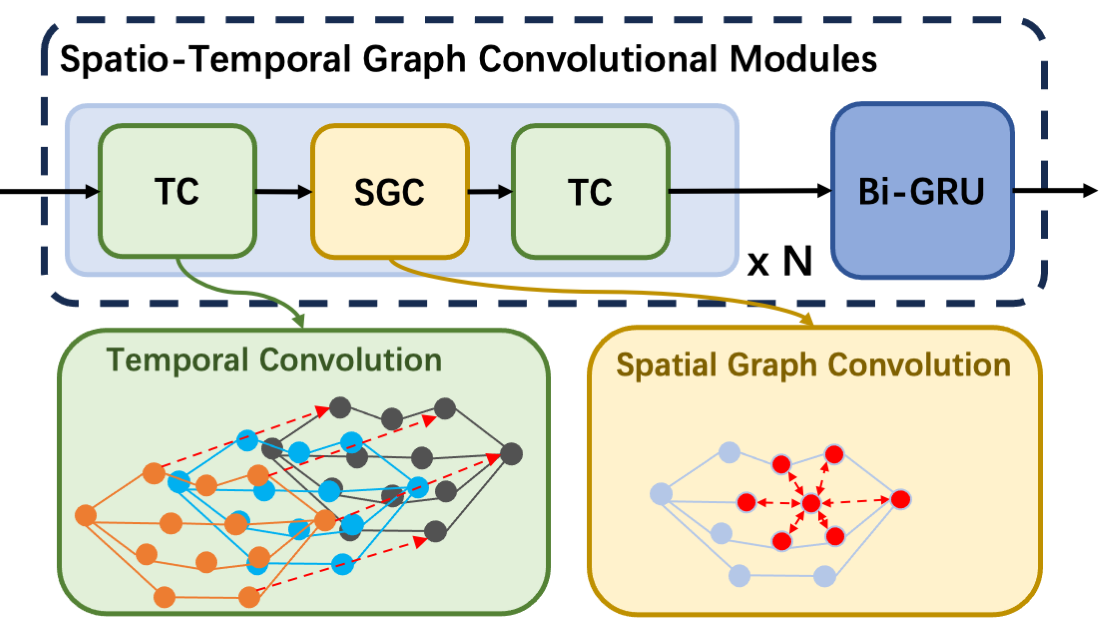}
    \caption{The overall architecture of spatio-temporal graph convolution module. Channel reduction and residual connection is ommited.}
    \label{fig:stgcn}
\end{figure}

The spatio-temporal graph convolutional modules capture dynamic changes and structural features in video samples through integrating information from both temporal and spatial dimensions, which consists of multiple spatio-temporal graph convolution layers followed by a final GRU module, as shown in Fig.~\ref{fig:stgcn}.

The spatio-temporal graph convolution layer consists of two temporal convolutions layers and one spatial graph convolution layer as in \cite{yu2018spatio}, follows ~(\ref{equ:stgcn-layer}). Specifically, the temporal convolution is implemented using 1D convolution, while the spatial graph convolution is achieved through time-distributed graph convolution.
\begin{equation}
    {\text{ST-GCN}}({\bf F}_{node}, {\bf M}_{adj})=  \text{TC}(\text{SGC}(\text{TC}({\bf F}_{node}), {\bf M}_{adj}))
    \label{equ:stgcn-layer}
\end{equation}
where the constructed node feature \({\bf F}_{node}\in \mathbb{R}^{B\times N\times T\times  C}\).

The purpose of temporal convolution is to capture local variations along the time dimension. By employing a smaller receptive field, temporal convolution can effectively capture short-term dependencies and local patterns within time series, enhancing the understanding of dynamic changes within time series. The parameter sharing mechanism allows the model to share the same convolution kernel when processing long sequences, thereby avoiding excessive computational complexity.

The spatial graph convolution is employed to capture structural features in the spatial dimension. The core idea of graph convolution is to define the connectivity between nodes through the adjacency matrix of the graph and utilize these connections to propagate and aggregate the feature information of nodes. The specific implementation of graph convolution follows ~(\ref{equ:gcn}). In this manner, graph convolutions can effectively capture spatial dependencies within graph structures and extract useful spatial features.
\begin{equation}
    \text{SGC}({\bf F}_{node}) = \sigma ({\bf M}_{adj}\times {\bf F}_{node}\times {\bf W})
    \label{equ:gcn}
\end{equation}
Here, \({\bf M}_{adj}\) represents the adjacency matrix, indicating the connections between nodes; \({\bf F}_{node}\) is the input node feature matrix; \({\bf W} \in \mathbb{R}^{C\times C}\) is the learnable weight matrix; and \(\sigma\) denotes the non-linear activation function.

At the end of the spatial-temporal graph convolutional module, a bidirectional GRU (Bi-GRU) is employed to synthesize global information. GRU is an improved version of the Recurrent Neural Network (RNN), offering stronger memory capabilities and faster training speeds compared to traditional RNNs. The bidirectional GRU further enhances the model's expressive power by considering both past and future information simultaneously.

The spatial-temporal graph convolutional module integrates temporal convolutions to capture local changes in the temporal dimension, graph convolutions to capture structural features in the spatial dimension, and bidirectional GRU to synthesize global information, achieving multi-level feature extraction and information fusion. This allows the model to effectively handle complex spatial-temporal sequence data.

\subsubsection{Lip Dynamics Fusion}
\label{sssec:fusion-explain}

Features from different landmark graph are fused and combined with the frame-wsie visual features before entering the model backend.
Different landmark graph focus on various motion patterns, and their fusion methods directly impact the model's understanding of temporal information and its classification capabilities.

A multi-level lip dynamic fusion framework is introduced.
Considering the similarity between DAG and SAG during construction, the features of DAG and SAG are concatenated with channel reduction. These combined features are integrated with the transformed LCG features. Finally, the feature from graph convolution module is fused with the frame-wise visual features.
The feature fusion pipeline will be discussed in Section~\ref{ssec:fusion}

\subsection{Temporal Feature Aggregation and Classification}
\label{ssec:tcn}

In the model backend, DC-TCN is employed to aggregate temporal information, which could effectively capture long-range dependencies in video sequences. 
DC-TCN expands the receptive field through dilated convolutions, enabling the model to capture information over longer time spans. This capability allows the model to better understand and analyze the complete dynamics of speech videos, effectively capturing temporal dependencies between segments. Compared to RNNs, DC-TCN allows for parallel computation, significantly reducing computational complexity.

A MLP-based classification head is employed to derive the spoken content from the aggregated features obtained from the backend. During training, the cross entropy loss with label smoothing is employed.
\begin{equation}
    L = - \left[(1 - \epsilon)\log(p_{true}) + \frac{\epsilon}{C-1}\sum_{i\neq true}  \log(p_i)\right]
\end{equation}

\section{Experiments}
\label{sec:experiments}

\subsection{Dataset}
\label{ssec:dataset}

\begin{table}[!t]
    \centering
    \caption{Comparison of lip dynamics fusion methods on LRW-ID Result under low-resource scenario}
    \label{tab:fusion}
    \begin{tabular}{lccc}
        \toprule
        Fusion                          & Acc &  mAcc & \#Params(M) \\
        \midrule
        \(\text{cat}(f_{lcg}, f_{sag})\) & 0.8623 & 0.8372 & 59.37 \\ 
        \(\text{sum}(f_{lcg}, f_{sag})\) & 0.8648 & 0.8385 & 58.72 \\
        \midrule
        \(\text{cat}(f_{lcg}, f_{dag})\) & 0.8638 & 0.8410 & 59.37 \\ 
        \(\text{sum}(f_{lcg}, f_{dag})\) & 0.8628 & 0.8402 & 58.72 \\
        \midrule
        \(\text{w-sum}(f_{lcg}, f_{dag}, f_{sag})\)          & 0.8545 & 0.8290 & 61.83 \\ 
        \(\text{cat}(f_{lcg}, f_{dag}, f_{sag})\)            & 0.8640 & 0.8404 & 62.90 \\ 
        \(\text{sum}(f_{lcg}, \text{cat}(f_{dag}, f_{sag})\) & 0.8665 & 0.8425 & 62.48 \\ 
        \bottomrule
    \end{tabular}
\end{table}

\begin{table*}[!t]
    \centering
    \caption{Ablation study on LRW-ID Result}
    \label{tab:ablation}
    \begin{tabular}{lcccc}
        \toprule
        Model & {Low-Resource Acc} & {High-Resource Acc}  & {\#Params (M)} & {\#FLOPS (G)} \\
        \midrule
        baseline         & 0.8575 & 0.9115  & 52.55 & 10.69 \\
        + DAG            & 0.8574 & 0.9138  & 55.65 & 10.82 \\
        + DAG + LCG       & 0.8628 & 0.9140  & 58.72 & 10.94 \\
        + DAG + LCG + SAG & \textbf{0.8665} & \textbf{0.9148} & 62.48 & 11.09 \\
        \bottomrule
    \end{tabular}
\end{table*}

The experiments are conducted over LRW and LRW-ID, both of which are commonly used in word-level visual speech recognition tasks. The details of these two datasets are described as follows.
\begin{itemize}
    \item LRW\cite{chung2017lip} is a publicly available dataset for lipreading of isolated words. The dataset is collected from more than 1000 speakers in BBC TV broadcasts. The number of target words is 500. Each video clip have 29 frames (1.16 seconds). The sizes of training, validation and test sets are 488766, 25000 and 25000 respectively.
    \item LRW-ID\cite{Kim2022} is a word-level lipreading dataset based on the LRW dataset, labelled with identity information. It consists of 500 English word classes.
    In user adaptive lip reading scenarios, 20 speakers (each with at least 900 videos) are selected to test the adaptation performance, and the others are used for training baseline models.
\end{itemize}

In the experimental process, the same settings as previous studies\cite{Kim2022} were followed.
Under the high-resource scenario, the entire training set from the original LRW-ID dataset was used for training, and the adaptation set from 20 selected speakers was used for validation and testing. 
In the low-resource scenario, one third of the original training set was selected for model training, while the complete adaptation set was used for validation and testing. 
When selecting the training set, videos were uniformly sampled from each identity ID in the training set to ensure balanced representation of all users during the training process.

\subsection{Implementation Details and Metrics}
\label{ssec:details}

Video frames of LRW dataset and LRW-ID dataset are cropped to 96 × 96.
During the training process, we employ random horizontal flipping and temporal masking techniques\cite{ma2022training} for data augment, but omit MixUp.
The visual feature extractor and back-end model structure is followed the setting in \cite{ma2022training}. 
The dimensionality of the extracted features is 512.
During training process, AdamW optimizer is adopted.
The cross-entropy loss incorporates label smoothing with a smoothing value of 0.1.
All the landamarks are extracted and localized from \cite{Martinez2020}, which based on open-source tools\footnote{online access: \text{https://github.com/hhj1897/face\_alignment}}.

Experiments are conducted with RTX2080Ti(11GB). In low-resource settings, only one third data of train set is utilized in train stage and the model is tested over all data from test set. The training time is ~4 days under low-resource scenario and ~9 days under high-resource scenario.
Our code will be made open source on \url{https://github.com/YangLeiSX/STMGCN}.

The overall classification accuracy (Acc) and the mean accuracy (mAcc) are employed as evaluation metrics. Here, \(M\) represents the number of users in the test set, \(N_u\) denotes the number of video samples from user \(u\), and \(N\) is the total number of samples in the test set.
\begin{align}
    Acc &= \frac{1}{N} \sum_{i}^{N} \mathbb{I}(\hat y_i == y_i)\\
    mAcc &= \frac{1}{M} \sum_{u=0}^{M} \frac{1}{N_u} \sum_{i}^{N_u} \mathbb{I}(\hat y_i^u == y_i^u)
\end{align}

\subsection{Lip Dynamics Fusion}
\label{ssec:fusion}

In the proposed Spatial-Temporal Multi-Graph Convolutional Module, there are three distinct graph. Features from different graphs are fused and subsequently combined with the origional visual features before being fed into the model's backend.

Different feature fusion methods are evaluated in the Spatial-Temporal Multi-Graph Convolutional Module and the results is shown as Table~\ref{tab:fusion}, where $cat$ means concatenate with channel reduction and $sum$ means point-wise addition. 
As for the $\text{w-sum}$ approach, a lightweight routing network computes the weights for different feature maps from each frame, merging them through a mixture-of-experts mechanism.
The experimental results indicate that while the concatenation-based feature fusion method can better preserve features from different branches, it does not significantly outperform the element-wise addition method. In contrast, the element-wise addition method incurs no additional parameters or computational costs.

The experimental results shown in Table~\ref{tab:fusion} also suggest that since the construction methods of DAG and SAG are quite similar, merging SAG and DAG before combining them with other features can effectively enhance the model's performance. In the following experiments, the $sum$ is adopted for two graph fusion and \(\text{sum}(f_{lcg}, \text{cat}(f_{dag}, f_{sag}))\) is adopted for three graph fusion.

\subsection{Ablation Study}
\label{ssec:ablation}

The experimental results of ablation study is shown as Table~\ref{tab:ablation}. 
The experimental results indicate that the distance-aware graph has a negligible impact on model performance in low-resource scenarios but can enhance performance in high-resource scenarios. This may be due to the fact that in images of the lip region, the distances between facial landmarks extracted by face recognition and localization models are relatively close. Consequently, the adjacency matrix of the DAG exhibits minimal variation in relative distances between different nodes, leading to over-smoothing during GCN processing and thereby failing to extract discriminative features effectively.

The incorporation of a landmark graph can significantly enhance model performance in low-resource scenarios, which aligns with our previous analysis that in resource-constrained settings, utilizing facial landmark information as a guide enables the model to more rapidly capture key visual features in the video, thereby improving performance. Additionally, since the facial landmark features are derived from the original video frames, as the amount of training data and training time increases, the model's performance tends to converge.

The incorporation of SAG, as opposed to LCG and DAG, offers complementary features by introducing similarity measures beyond lip contour and landmark distances. In videos, regions with similar features often belong to the same semantic categories and serve similar functions. This mirrors the human process of observing speech videos, where one first identifies rough regions such as the face, lips, and various oral components based on color and texture similarity, followed by the observation of motion dynamics across these regions. The addition of SAG has led to improved classification performance across both low-resource and high-resource scenarios.

\subsection{Comparison with Other Methods}
\label{ssec:sota}

\begin{table}[!t]
  \centering
  \caption{Performance comparisons on LRW-ID dataset}
  \label{tab:comparison}
  \begin{tabular}{lcccc}
      \toprule
      \multirow{2}{*}{Model}                         & \multicolumn{2}{c}{Low-Resource}  & \multicolumn{2}{c}{High-Resource} \\
      & Acc & mAcc & Acc & mAcc \\
      \midrule
      UDP\cite{kim2022speaker}    & ---    & ---    & ---    & 0.8585 \\
      He's\cite{he2024speaker}    & ---    & ---    & ---    & 0.8848 \\
      DC-TCN\cite{ma2022training} & 0.8575 & 0.8341 & 0.9115 & 0.8944 \\
      \midrule
      LipFormer\cite{Xue2023}     & 0.8187 & 0.7862 & 0.8872 &   0.8641 \\
      GusLip\cite{Li2024}         & 0.8407 & 0.8100 & 0.8984 &    0.8763 \\
      \midrule
      Proposed                    & \textbf{0.8665} & \textbf{0.8425} & \textbf{0.9148} & \textbf{0.8961} \\
      \bottomrule
  \end{tabular}
\end{table}

\begin{table}[!t]
  \centering
  \caption{Performance comparisons on LRW dataset }
  \label{tab:comparison-lrw}
  \begin{tabular}{lcc}
      \toprule
      Model &  Acc \\
      \midrule
      TSM \cite{hao2021use} & 0.8623 \\ 
      VTM \cite{sheng2022importance} & 0.8692 \\ 
      DC-TCN \cite{ma2022training} & 0.9100$^a$ \\ 
      \midrule
      LipFormer \cite{Xue2023} & 0.8691 \\ 
      GusLip \cite{Li2024} & 0.8915 \\ 
      \midrule
      Proposed & \textbf{0.9169} \\ 
      \bottomrule
  \end{tabular}

  \vspace{0.5em}
  \footnotesize{$^a$: without mixup.}
\end{table}

The comparison results with other word-level lipreading models on the LRW-ID dataset are shown in Table~\ref{tab:comparison}. 
In contrast to models that rely exclusively on video features, the additional facial landmark features provide the model with richer information, leading to a noticeable improvement in performance. Even without using user data for adaptive fine-tuning, better results can still be achieved on unseen speakers.

In contrast to models that use facial landmarks to construct graph convolutions, our approach leverages spatio-temporal multi-graph convolutions, fully utilizing the coordinates of the facial landmark sequences, the similarity between different features, and the distances between different landmarks, thereby maximizing the utilization of landmark information.

Comparative experiments were conducted on the complete LRW dataset, and the model's recognition accuracy is shown in the Table~\ref{tab:comparison-lrw}. It can be observed that leveraging landmark information to construct spatio-temporal multi-graph convolutional networks significantly enhances the model's recognition efficiency.

\subsection{Robustness Analysis}
\label{ssec:robust}

\begin{table}[t]
  \centering
  \caption{The Robustness Test}
  \label{tab:robust}
  \begin{tabular}{lcccc}
  \toprule
  \textbf{Model} & \multicolumn{2}{c}{\textbf{Low-Resource}} & \multicolumn{2}{c}{\textbf{High-Resource}} \\
   & Acc & mAcc & Acc & mAcc \\
  \midrule
  DC-TCN\cite{ma2022training}  & 0.8575 & 0.8341 & 0.9115 & 0.8944 \\
  + Noise + Pertubation & 0.8479 & 0.8237 & 0.8992 & 0.8796 \\
  \midrule
  LipFormer\cite{Xue2023} & 0.8187 & 0.7862 & 0.8872 & 0.8641 \\
  + Noise + Pertubation & 0.8015 & 0.7660 & 0.8684 & 0.8442 \\
  \midrule
  GusLip\cite{Li2024}  & 0.8407 & 0.8100 & 0.8984 & 0.8763 \\
  + Noise + Pertubation & 0.8329 & 0.8024 & 0.8943 & 0.8708 \\
  \midrule
  Proposed & 0.8665 & 0.8425 & 0.9148 & 0.8961 \\
  + Visual Noise & 0.8579 & 0.8343 & 0.9071 & 0.8882 \\
  + Landmark Perturbation & 0.8645 & 0.8416 & 0.9146 & 0.8959 \\
  + Noise + Pertubation & 0.8580 & 0.8333 & 0.9075 & 0.8875 \\
  \bottomrule
  \end{tabular}
\end{table}

We have conducted experiments to evaluate the robustness of the proposed method to random perturbations in the facial landmarks and noise in image.
The results of the robustness test on LRW-ID dataset are shown in Table \ref{tab:robust}. We specifically examined two types of disturbances: visual noise and landmark perturbations, the former is conducted by adding Gaussian noise to the input images, and the latter is implemented by introducing small coordinate offsets to the facial landmarks. This approach allows us to evaluate how minor shifts in landmark locations impact the model's performance. The results show that the proposed method is robust to random perturbations in the facial landmarks and facial image noise. 

The reasons are multifaceted. Firstly, the random data augmentation techniques are employed on the input images, enhancing the model robustness against variations and perturbations. Secondly, the proposed method based on multi-graph convolutional networks is designed to fusion information from multiple perspectives such as landmark coordinates trajectory, spatial distance and semantic similarity of spatio-temporal features, which possess robustness to landmark perturbations and visual noise.

\subsection{Discussion}
\label{ssec:disc}

Considering that the preprocessing of raw video in visual speech recognition systems already includes face detection, face alignment, and lip region cropping, all of which rely on the results of facial landmark detection, the proposed method does not incur additional computational costs for extracting landmark positions when integrated into existing systems.
However, the model proposed in this paper still has certain limitations. The spatial-temporal graph convolution employs matrix multiplication, which results in a large number of model parameters and computational overhead. This may hinder the deployment and utilization of the model on edge computing devices. In future works, it is advisable to explore the use of sparse matrices and sparse computation techniques, particularly for the computation of LCG, to enhance computational efficiency.

\section{Conclusion}
\label{sec:conclusion}

This paper introduces a landmark guided visual feature extractor for visual speech recognition. Three distinct landmark graphs, Landmark Coordinate Graph (LCG), Distance-Aware Graph (DAG), and Similarity-Aware Graph (SAG), are constructed and integrated via a spatio-temporal multi-graph convolutional module. Experimental results demonstrate that the proposed method significantly improves the model's performance on unseen speakers, even with limited training data.

\bibliographystyle{ACM-Reference-Format}
\balance
\bibliography{main}

\end{document}